\documentclass{article}

\usepackage{graphicx}
\usepackage{pstricks}

\usepackage{nemlap3}

\title{
Look-Back and Look-Ahead in the Conversion of \\
Hidden Markov Models into Finite State Transducers}

\author{
Andr\'e Kempe \vspace{1mm} \\
{\normalsize Xerox Research Centre Europe -- Grenoble Laboratory} \\
{\normalsize 6, chemin de Maupertuis -- 38240 Meylan -- France}\vspace{1mm}\\
{\normalsize \tt andre.kempe@xrce.xerox.com} \\
{\normalsize \tt http://www.xrce.xerox.com/research/mltt} }

\begin{document}

\newcommand{\mycomment}[1]
{ \rule{0mm}{0mm} \\
\begin{center}
\fbox{
\begin{minipage}{75mm}
\noindent \underline{\large \bf \it COMMENT :} \\

\vspace{-2mm}
 #1 
\end{minipage} }
\end{center}
 \rule{0mm}{0mm} \\ }

\newcounter{NumRE}     

\newcommand{\REa}[2]           
{ \hfill
\refstepcounter{NumRE}
\makebox[120mm][#1]{{\tt #2}}  
\makebox[15mm][r]{[\theNumRE]} \\ } 

\newcommand{\REaL}[3]          
{ \REa{#1}{#2} \label{#3} }    

\newcommand{\REb}[1]           
{ \hfill
\refstepcounter{NumRE}
\parbox[t]{120mm}{{\tt #1}}                 
\parbox[t]{15mm}{\hfill [\theNumRE]} \\ }   

\newcommand{\REbL}[2]          
{ \REb{#1} \label{#2} }

\newcommand{\EQb}[1]           
{ \hfill
\refstepcounter{NumRE}
\parbox[t]{120mm}{{${#1}$}}                 
\parbox[t]{15mm}{\hfill [\theNumRE]} \\ }

\newcommand{\eqcom}[2]         
{\hfill \begin{minipage}{#1} {\it #2} \end{minipage} }

\newcommand{\RStep}[5]         
{\noindent (#1) #2 \\          
\vspace{-3mm} \\

\REbL{#4}{#3}                  
\vspace{-3mm} \\

\eqcom{140mm}{#5} }

\newcommand{\otherfootnote}[1]  
{$^{\ref{#1}}$}

\newcommand{\LeftBr}[1]{ {\tt \LARGE \bf <}$_{#1}$}
\newcommand{\RightBr}[1]{ {\tt \LARGE \bf >}$_{#1}$}

\def\contains{\$}               
\def\xnot{\tilde{\enspace}}
\def\notcont{\xnot \contains}
\def\ntimes{\hat{\enspace}}
\def\xprime{\!\acute{\enspace}}

\def\ctxU{$||$}
\def\ctxR{$//$}
\def\ctxL{$\backslash \backslash$~}
\def\ctxD{$\backslash /$~}

\def\igin{\tt .\rule{-0.8mm}{0mm}/\rule{-1.6mm}{0mm}.}

\def\eps{\epsilon}
\def\bs{\backslash}
\def\SmSpc{\rule{1.5mm}{0mm}}
\def\cct{$^{\frown}$}
\def\CtxBd{{\tt.\#.}}

\def\kleenestar{\!\!*}
\def\kleeneplus{\!{\tt+}}
\def\crosspr{\; {\tt .x.}\; }
\def\compose{\; {\tt .o.}\; }

\def\repl{\; {\tt -\!\!\!>}\; }
\def\invrepl{\; {\tt <\!\!\!-}\; }
\def\birepl{\; {\tt <\!\!-\!\!\!>}\; }
\def\symp{\!:\!}
\def\req{\!\!-\!\!}
\newcommand{\OLPair}[2]{ \langle #1,#2 \rangle }
\def\OneL{.{\it1L}}
\def\TwoL{.{\it2L}}
\newcommand{\Xlevel}[1]{\overline{#1}}

\def\longbar{\rule[1mm]{9mm}{0.2mm}}
\def\ambr{$^*$}
\def\nac{\rule[-0.5mm]{9mm}{0.2mm}}
\def\nag{ }

\newcommand{\lxu}[1]{\rule{-2mm}{0mm}\enspace^{#1}\rule{-0.5mm}{0mm}} 
\newcommand{\lxl}[1]{\rule{-2mm}{0mm}\enspace_{#1}\rule{-0.5mm}{0mm}} 
\newcommand{\lxul}[2]{\rule{-2mm}{0mm}\enspace_{#2}^{#1}\rule{-0.5mm}{0mm}}

\newcommand{\spc}[1]{\rule{#1}{0mm}}
\newcommand{\vspc}[1]{\rule{0mm}{#1}}

\def\SmSpc{\spc{1.5mm}}
\def\spA{\spc{3mm}}
\def\spB{\spc{6mm}}

\def\SpcUp{\rule{0mm}{4.5mm}}
\def\SpcDown{\rule[-1mm]{0mm}{2mm}}

\def\mybibbegin{\begin{minipage}{79mm}
                \hfill \begin{minipage}{76mm}}

\def\mybibend{\end{minipage} \end{minipage}}

\def\mybibitem{\vspace{1mm}
               \noindent \rule{-3mm}{0mm}}

\maketitle

\begin{abstract}
This paper describes the conversion of a Hidden Markov Model
into a finite state transducer
that closely approximates the behavior of the stochastic model.
In some cases the transducer is equivalent to the HMM.
This conversion is especially advantageous for 
part-of-speech tagging
because the resulting transducer can be composed with 
other transducers that encode correction rules
for the most frequent tagging errors.
The speed of tagging is also improved.
The described methods have been implemented and
successfully tested.
\end{abstract}

\section{Introduction \label{s-intro}}

This paper presents an algorithm\footnote{
There are other (different) algorithms for HMM to FST conversion:
An unpublished one by Julian M. \mbox{Kupiec} and John T. Maxwell (p.c.),
and n-type and s-type approximation by Kempe (1997).
}
which approximates a {\it Hidden Markov Model} (HMM)
by a {\it finite-state transducer} (FST).
We describe one application, namely part-of-speech tagging.
Other potential applications may be found in areas
where both HMMs and finite-state technology are applied,
such as speech recognition, etc.
The algorithm has been fully implemented.

An HMM used for tagging encodes, like a transducer,
a relation between two languages.
One language contains sequences of ambiguity classes
obtained by looking up in a lexicon all words of a sentence.
The other language contains sequences of tags
obtained by statistically disambiguating the class sequences.
From the outside, an HMM tagger behaves like a sequential transducer
that deterministically maps every class sequence to a tag sequence,
e.g.:
\begin{equation}
\frac{\tt{[DET,PRO] ~[ADJ,NOUN] ~[ADJ,NOUN] ~...... ~[END]}}
 {\tt{\spc{5mm}DET\spc{11mm}ADJ\spc{11mm}NOUN\spc{3mm} ~......\spc{2mm}END}}
  \label{e-CtoT}
\end{equation} \\

The main advantage of  transforming an HMM is that the resulting
transducer can be handled by finite state calculus.
Among others, it can be composed with transducers that encode:
\begin{itemize}
\vspace{-1.5mm}
\item
correction rules for the most frequent tagging errors
which are automatically generated
(Brill, 1992; Roche and Schabes, 1995)
or manually written (Chanod and Tapanainen, 1995),
in order to significantly improve tagging accuracy\footnote{
Automatically derived rules require less work than manually written ones
but are unlikely to yield better results because they would consider
relatively limited context and simple relations only.
}.
These rules may include
long-distance dependencies not handled by HMM taggers,
and can conveniently be expressed by the replace operator
(Kaplan and Kay, 1994; Karttunen, 1995; Kempe and Karttunen, 1996).
\vspace{-1.5mm}
\item
further steps of text analysis, e.g. light parsing or extraction
of noun phrases or other phrases (A\"{\i}t-Mokhtar and Chanod, 1997).
\end{itemize}
\vspace{-1.5mm}

\noindent
These compositions enable complex text analysis to be performed
by a single transducer.

The speed of tagging by an FST is up to six times higher
than with the original HMM.

The motivation for deriving the FST from an HMM is that the HMM
can be trained and converted with little manual effort.

An HMM transducer builds on the data (probability matrices)
of the underlying HMM.
The accuracy of this data
has an impact on the tagging accuracy of both the HMM itself
and the derived transducer.
The training of the HMM can be done
on either a tagged or untagged corpus, and is not a topic
of this paper since it is exhaustively described in the literature
(Bahl and Mercer, 1976; Church, 1988).

An HMM can be identically represented by a weighted FST
in a straightforward way.
We are, however, interested in non-weighted transducers.

\section{b-Type Approximation \label{s-btype}}

This section presents a method that approximates a (first order)
Hidden Markov Model (HMM) by a finite-state transducer (FST),
called {\it b-type} approximation\footnote{
Name given by the author, to distinguish the algorithm
from n-type and s-type approximation (Kempe, 1997).
}. Regular expression operators used in this section
are explained in the annex.

Looking up, in a lexicon, the word sequence of a sentence
produces a unique sequence of ambiguity classes.
Tagging the sentence by means of a (first order) HMM
consists of finding the most probable tag sequence $T$
given this class sequence $C$
(eq.~\ref{e-CtoT}, fig.~\ref{f-bseq-b}).
The joint probability of the sequences $C$ and $T$
can be estimated by:
\begin{eqnarray}
p (C,T) = p (c_1 .... c_n , t_1 .... t_n) =  \spc{15mm} \nonumber \\
   \spc{10mm} \pi (t_1) \; b (c_1|t_1) \cdot
       \prod\limits_{i=2}^{n} a (t_i|t_{i-1}) \; b (c_i|t_i)
\end{eqnarray}

\vspace{3mm}
\subsection{Basic Idea}
\vspace{1mm}

The determination of a tag of a particular word cannot be made
separately from the other tags.
Tags can influence each other over a long distance via
transition probabilities.

In this approach,
an ambiguity class is disambiguated with respect to a context.
A context consists of a sequence of ambiguity classes
limited at both ends by some selected tag\footnote{
The algorithm is explained for a first order HMM.
In the case of a second order HMM, b-type sequences
must begin and end with two selected tags rather than one.
  \label{n-sec-hmm}
}.
For the left context of length $\beta$ we use the term {\it look-back},
and for the right context of length $\alpha$ we use the term {\it look-ahead}.

\begin{figure}[htb]
\begin{center}
\includegraphics[scale=0.5,angle=0]{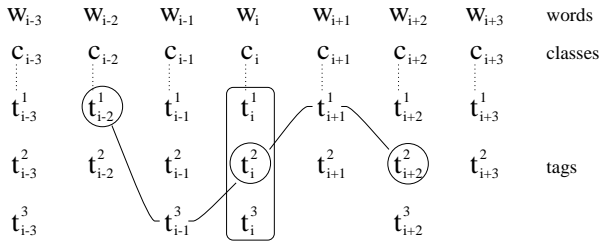}
\begin{minipage}{70mm}
\caption{Disambiguation of classes between two selected tags
    \label{f-bseq-b}}
\end{minipage}
\end{center}
\end{figure}

In figure \ref{f-bseq-b}, the tag $t_i^2$ can be selected from
the class $c_i$ because it is between two selected
tags\otherfootnote{n-sec-hmm}
which are $t_{i-2}^1$ at a look-back distance of $\beta=2$
and $t_{i+2}^2$ at a look-ahead distance of $\alpha=2$.
Actually, the two selected tags $t_{i-2}^1$ and $t_{i+2}^2$ allow
not only the disambiguation of the class $c_i$ but of all classes
inbetween, i.e. $c_{i-1}$, $c_i$ and $c_{i+1}$.

\begin{figure}[htbp]
\begin{center}
\scalebox{1 1.4}{\includegraphics[scale=0.48,angle=0]{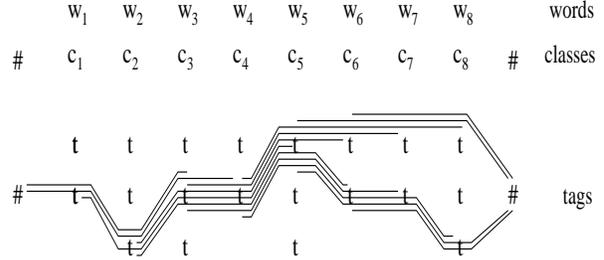}}
\begin{minipage}{70mm}
\caption{Two valid paths through the tag space of a sentence
    \label{f-bpath-val}}
\end{minipage}
\end{center}
\end{figure}

We approximate the tagging of a whole sentence by tagging subsequences
with selected tags at both ends (fig.~\ref{f-bseq-b}),
and then overlapping them.
The most probable paths in the tag space of a sentence, i.e. valid paths
according to this approach, can be found as sketched in figure
\ref{f-bpath-val}.

\begin{figure}[htbp]
\begin{center}
\scalebox{1 1.4}{\includegraphics[scale=0.48,angle=0]{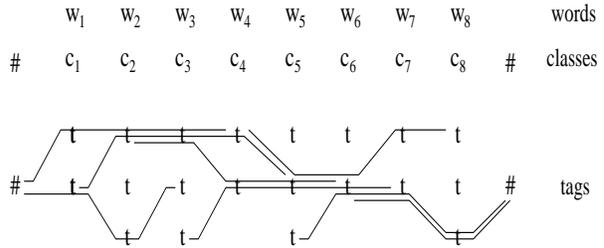}}
\begin{minipage}{70mm}
\caption{Incompatible sequences in the tag space of a sentence
    \label{f-bpath-inval}}
\end{minipage}
\end{center}
\end{figure}

A valid path consists of an ordered set of overlapping sequences in which
each member overlaps with its neighbour except for the first or last tag.
There can be more than one valid path
in the tag space of a sentence (fig.~\ref{f-bpath-val}).
Sets of sequences that do not overlap in such a way are incompatible
according to this model, and do not constitute valid paths
(fig.~\ref{f-bpath-inval}).

\vspace{3mm}
\subsection{b-Type Sequences}
\vspace{1mm}

Given a length $\beta$ of look-back and a length $\alpha$ of look-ahead,
we generate for every class $c_0$,
every look-back sequence $t_{-\beta} \; c_{-\beta+1} \; ... \; c_{-1}$,
and every look-ahead sequence $c_{1} \; ... \; c_{\alpha-1} \; t_{\alpha}$,
a b-type sequence\otherfootnote{n-sec-hmm}:
\begin{equation}
t_{-\beta} \; c_{-\beta+1} \; ... \; c_{-1} \; \;
c_0 \; \;
c_{1} \; ... \; c_{\alpha-1} \; t_{\alpha}
 \label{e-origBSeq}
\end{equation}

\pagebreak
\noindent
For example:
\begin{equation} \scalebox{0.9 1}{
{\tt
CONJ  \;
[DET,PRON]  \;
[ADJ,NOUN,VERB]  \;
[NOUN,VERB]  \;
VERB
} } \label{e-bseq-exm0}
\end{equation}

\begin{figure}[htbp]
\begin{center}
\includegraphics[scale=0.5,angle=0]{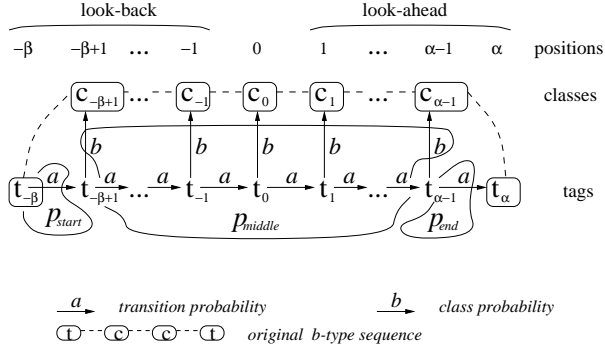}
\begin{minipage}{70mm}
\caption{b-Type sequence
    \label{f-bseq-c}}
\end{minipage}
\end{center}
\end{figure}

Each such {\it original b-type sequence}
(eq.~\ref{e-origBSeq},\ref{e-bseq-exm0}; fig.~\ref{f-bseq-c})
is disambiguated based on a first order HMM.
Here we use the Viterbi algorithm (Viterbi, 1967; Rabiner, 1990)
for efficiency.

For an original b-type sequence,
the joint probability of its class sequence $C$ with its tag sequence $T$
(fig.~\ref{f-bseq-c}), can be estimated by:
\begin{eqnarray}
p (C,T) =
p (c_{-\beta+1} \; ... \; c_{\alpha-1} \; ,
    \; t_{-\beta} \; ... \; t_{\alpha}) =     \spc{6mm} \nonumber \\
\left[ \prod\limits_{i=-\beta+1}^{\alpha-1} \spc{-2mm}
          \; a (t_i | t_{i-1}) \; b (c_i | t_i) \right]
 \, \cdot \, a (t_{\alpha} | t_{\alpha-1})
  \label{e-Pr-bseq}
\end{eqnarray}

At every position in the look-back sequence and in the look-ahead sequence,
a boundary \# may occur, i.e. a sentence beginning or end.
No look-back ($\beta\!=\!0$) or no look-ahead ($\alpha\!=\!0$)
is also allowed.
The above probability estimation (eq.~\ref{e-Pr-bseq})
can then be expressed more generally (fig.~\ref{f-bseq-c}) as:
\begin{equation}
p (C,T) = p_{start} \cdot p_{middle} \cdot p_{end}
  \label{e-Pr-bseq-gen}
\end{equation}

\noindent
with $p_{start}$ being
\begin{eqnarray}
p_{start} = &\spc{-3mm} a (t_{-\beta+1}|t_{-\beta})
                         \spc{0mm} & {\rm for~selected~tag} \; t_{-\beta}\spc{4mm} \\
p_{start} = &\spc{-3mm} \pi (t_{-\beta+1}) \spc{7mm} & {\rm for~boundary}\; \# \\
p_{start} = &\spc{-3mm} 1                  \spc{15mm} & {\rm for}\; \beta\!=\!0
\end{eqnarray}

\noindent
with $p_{middle}$ being
\begin{eqnarray}
p_{middle} = &\spc{-3mm} b (c_{-\beta+1} | t_{-\beta+1}) \cdot &\spc{-4mm}
   \prod\limits_{i=-\beta+2}^{\alpha-1}\spc{-3mm} a (t_i|t_{i-1})\; b (c_i | t_i)
         \nonumber \\
            & \spc{0mm} &\spc{6mm} {\rm for} \; \alpha\!+\!\beta\!>\!0\spc{0mm} \\
p_{middle} = &\spc{-3mm} b (c_{0} | t_{0})
           \spc{10mm} & {\rm for} \; \alpha\!+\!\beta\!=\!0
\end{eqnarray}

\noindent
and with $p_{end}$ being
\begin{eqnarray}
p_{end} = &\spc{-3mm} a (t_{\alpha}|t_{\alpha-1})
                         \spc{0mm} & {\rm for~selected~tag} \; t_{\alpha} \\
p_{end} = &\spc{-3mm} 1  \spc{10mm} & {\rm for~boundary~\#~or} \; \alpha\!=\!0\spc{4mm}
  \label{e-PrEnd}
\end{eqnarray}

When the most likely tag sequence is found for an original
b-type sequence, the class $c_0$ in the middle position
(eq.~\ref{e-origBSeq}) is associated with its most likely tag $t_0$.
We formulate constraints for the other tags $t_{-\beta}$ and $t_{\alpha}$
and classes $c_{-\beta+1}...c_{-1}$ and $c_{1}...c_{\alpha-1}$
of the original b-type sequence.
Thus we obtain a {\it tagged b-type sequence}~\footnote{
Regular expression operators used in this article
are explained in the annex. \label{n-regex}
}:
\begin{equation} \scalebox{0.85 1}{
t_{-\beta}^{B\beta} \; c_{-\beta+1}^{B(\beta-1)}
        \; ... c_{-2}^{B2} \; c_{-1}^{B1} \; \;
c_{0} \symp t_{0} \; \;
c_{1}^{A1} \; c_{2}^{A2}
        \; ... c_{\alpha-1}^{A(\alpha-1)} \; t_{\alpha}^{A\alpha} }
  \label{e-tgdBSeq}
\end{equation}

\noindent
stating that $t_0$ is the most probable tag in the class $c_0$
if it is preceded by $t^{B\beta}\; c^{B(\beta-1)}...c^{B2}\; c^{B1}$
and followed by $c^{A1}\; c^{A2}...c^{A(\alpha-1)}\; t^{A\alpha}$.

In expression \ref{e-tgdBSeq} the subscripts

\scalebox{0.95 1}{$-\!\beta\, -\!\!\beta\!+\!\!1...0...\alpha\!\!-\!\!1$}
  $\alpha$
denote the position of the tag or class in the b-type sequence,
and the superscripts
$B\beta\; B(\beta\!\!-\!\!1)... B1$ and $A1...A(\alpha\!-\!1)\; A\alpha$
express constraints for preceding and following tags and classes
which are part of other b-type sequences.
In the example\otherfootnote{n-regex}:
\begin{eqnarray}
\scalebox{0.9 1}{\tt CONJ \req B2 \;\; [DET,PRON] \req B1}\spc{35mm}\nonumber\\
\scalebox{0.9 1}{\tt [ADJ,NOUN,VERB] \symp ADJ}  \spc{20mm}  \nonumber \\
\spc{30mm} \scalebox{0.9 1}{\tt [NOUN,VERB] \req A1  \;\; VERB \req A2}
  \label{e-bseq-exm1}
\end{eqnarray}

\noindent
{\tt ADJ} is the most likely tag in the class {\tt [ADJ,NOUN,VERB]}
if it is preceded by the tag {\tt CONJ} two positions back ({\tt B2}),
by the class {\tt [DET,PRON]} one position back ({\tt B1}),
and followed by the class {\tt [NOUN,VERB]} one position ahead ({\tt A1})
and by the tag {\tt VERB} two positions ahead ({\tt A2}).

Boundaries are denoted by a particular symbol \#
and can occur at the edge of the look-back and look-ahead sequence:
\small
\begin{eqnarray}
t^{B\beta} \; c^{B(\beta-1)} \; ... c^{B2} \; c^{B1}
  &\spc{-2mm}  c \symp t  &\spc{-2mm}
    c^{A1} \; c^{A1} \; ... c^{A(\alpha-1)} \; \#^{A\alpha}\spc{5mm}  \\
t^{B\beta} \; c^{B(\beta-1)} \; ... c^{B2} \; c^{B1}
  &\spc{-2mm}  c \symp t  &\spc{-2mm}
    c^{A1} \; c^{A1} \; ... \#^{A(\alpha-1)}  \\
\#^{B\beta} \; c^{B(\beta-1)} \; ... c^{B2} \; c^{B1}
  &\spc{-2mm}  c \symp t  &\spc{-2mm}
    \#^{A1}  \\
\#^{B1}
  &\spc{-2mm}  c \symp t  &\spc{-2mm}
    \#^{A1}  \\
\#^{B2} \; c^{B1}
  &\spc{-2mm}  c \symp t  &\spc{-2mm}
    c^{A1} \; c^{A1} \; ... c^{A(\alpha-1)} \; t^{A\alpha}
\end{eqnarray}
\normalsize

\noindent
For example:
\begin{eqnarray}
\scalebox{0.9 1}{\tt \# \req B2 \;\; [DET,PRON] \req B1}\spc{35mm}\nonumber \\
   \scalebox{0.9 1}{\tt [ADJ,NOUN,VERB] \symp ADJ}   \spc{20mm} \nonumber \\
\spc{30mm} \scalebox{0.9 1}{\tt [NOUN,VERB] \req A1  \;\; VERB \req A2}
  \label{e-bseq-exm2}
\end{eqnarray}

\pagebreak

\begin{eqnarray}
\scalebox{0.9 1}{\tt CONJ \req B2 \;\; [DET,PRON] \req B1}\spc{35mm}\nonumber\\
   \scalebox{0.9 1}{\tt [ADJ,NOUN,VERB] \symp NOUN}  \spc{18mm} \nonumber \\
\spc{30mm} \scalebox{0.9 1}{\tt \# \req A1} \spc{18mm}\enspace
  \label{e-bseq-exm3}
\end{eqnarray}

Note that look-back of length $\beta$ and look-ahead of length $\alpha$
also include all sequences shorter than $\beta$ or $\alpha$, respectively,
that are limited by \#.

For a given length $\beta$ of look-back and a length $\alpha$ of look-ahead,
we generate every possible original b-type sequence (eq.~\ref{e-origBSeq}),
disambiguate it statistically (eq.~\ref{e-Pr-bseq}-\ref{e-PrEnd}),
and encode the tagged b-type sequence $B_i$ (eq.~\ref{e-tgdBSeq})
as an FST.
All sequences $B_i$ are then unioned
\begin{equation}
\lxu{\cup}B = \bigcup\limits_i B_i
  \label{e-Bunion}
\end{equation}

\noindent
and we generate a preliminary tagger model $B\xprime$
\begin{equation}
B\xprime = [\; \lxu{\cup}B \;]\; \kleenestar
  \label{e-Bprime}
\end{equation}

\noindent
where all sequences $B_i$ can occur in any order and number
(including zero times) because no constraints have yet been applied.

\vspace{2mm}
\subsection{Concatenation Constraints \label{ss-cons}}
\vspace{1mm}

To ensure a correct concatenation of sequences $B_i$, we have to make
sure that every $B_i$ is preceded and followed by other $B_i$
according to what is encoded in the look-back and look-ahead constraints.
E.g. the sequence in example (\ref{e-bseq-exm2})
must be preceded by a sentence beginning, \#, and the class {\tt [DET,PRON]}
and followed by the class {\tt [NOUN,VERB]} and the tag {\tt VERB}.

We create constraints for preceding and following tags, classes
and sentence boundaries.
For the look-back, a particular tag $t_i$ or class $c_j$
is required for a particular distance of $\delta \le -1$,
by\otherfootnote{n-regex}:
\small
\begin{eqnarray}
R^{\delta} (t_i) &\spc{-5mm} = &\spc{-5mm}
 \xnot[\,
  \xnot[\, ?\kleenestar\; t_i\;
      [\bs\!\lxu{\cup}t]\kleenestar\;
      [\lxu{\cup}t \; [\bs\!\lxu{\cup}t]\;\kleenestar ]\ntimes{\scalebox{0.6 1}{(-\delta\!-\!1)}}\, ]
   \; t_i^{\scalebox{0.6 0.8}{B (-\delta)}}\; ?\;\kleenestar\, ]\spc{8mm}
  \label{e-tagConsB}  \\
\vspc{7mm}
R^{\delta} (c_j) &\spc{-5mm} = &\spc{-5mm}
 \xnot[\,
  \xnot[\, ?\kleenestar\; c_j\;
      [\bs\!\lxu{\cup}c]\kleenestar\;
      [\lxu{\cup}c \; [\bs\!\lxu{\cup}c]\;\kleenestar ]\ntimes{\scalebox{0.6 1}{(-\delta\!-\!1)}}\, ]
   \; c_j^{\scalebox{0.6 0.8}{B (-\delta)}}\; ?\;\kleenestar\, ]
  \label{e-classConsB}  \\
\vspc{6mm}
 & \enspace & \spc{32mm} {\rm for} \spc{4mm} \delta \le -1 \nonumber
\end{eqnarray}
\normalsize

\noindent
with $\lxu{\cup}t$ and $\lxu{\cup}c$ being
the union of all tags and all classes respectively.

A sentence beginning, \#, is required for a particular look-back
distance of $\delta\!\le\!-1$, on the side of the tags, by:
\small
\begin{eqnarray}
R^{\delta} (\#) &\spc{-5mm} = &\spc{-5mm}
  \xnot[\;
   \xnot[\;
        [\bs\!\lxu{\cup}t]\kleenestar\;
        [\lxu{\cup}t \; [\bs\!\lxu{\cup}t]\;\kleenestar ]\ntimes{\scalebox{0.6 1}{(-\delta\!-\!1)}}\, ]
   \; \#^{\scalebox{0.6 0.8}{B (-\delta)}}\; ?\;\kleenestar\, ]\spc{10mm}
  \label{e-bndConsB}  \\
\vspc{6mm}
 & \enspace & \spc{30mm} {\rm for} \spc{4mm} \delta \le -1 \nonumber
\end{eqnarray}
\normalsize

\pagebreak

In the case of look-ahead we require for a particular distance
of $\delta\!\ge\!1$, a particular tag $t_i$ or class $c_j$
or a sentence end, \#, on the side of the tags,
in a similar way by:
\small
\begin{eqnarray}
R^{\delta} (t_i) &\spc{-5mm} = &\spc{-5mm}
  \xnot[\, ?\kleenestar\; t_i^{\scalebox{0.6 0.8}{A\delta}}\;
    \xnot[\; [\bs\!\lxu{\cup}t]\kleenestar\;
         [\lxu{\cup}t \; [\bs\!\lxu{\cup}t]\;\kleenestar ]\ntimes{\scalebox{0.6 1}{(\delta-1)}}
     \; t_i\; ?\;\kleenestar\, ]\, ]\spc{10mm}
  \label{e-tagConsA}  \\
\vspc{7mm}
R^{\delta} (c_j) &\spc{-5mm} = &\spc{-5mm}
  \xnot[\; ?\kleenestar\; c_j^{\scalebox{0.6 0.8}{A\delta}}\;
    \xnot[\; [\bs\!\lxu{\cup}c]\kleenestar\;
         [\lxu{\cup}c \; [\bs\!\lxu{\cup}c]\;\kleenestar ]\ntimes{\scalebox{0.6 1}{(\delta-1)}}
     \; c_j\; ?\;\kleenestar\, ]\, ]
  \label{e-classConsA}  \\
\vspc{7mm}
R^{\delta} (\#) &\spc{-5mm} = &\spc{-5mm}
  \xnot[\; ?\kleenestar\; \#^{\scalebox{0.6 0.8}{A\delta}}\;
    \xnot[\; [\bs\!\lxu{\cup}t]\kleenestar\;
         [\lxu{\cup}t \; [\bs\!\lxu{\cup}t]\;\kleenestar ]\ntimes{\scalebox{0.6 1}{(\delta-1)}}
     \, ]\, ]
  \label{e-bndConsA}  \\
\vspc{6mm}
 & \enspace & \spc{31mm} {\rm for} \spc{4mm} \delta \ge 1 \nonumber
\end{eqnarray}
\normalsize

All tags $t_i$ are required for the look-back only
at the distance of $\delta = -\beta$
and for the look-ahead only at the distance of $\delta\!=\!\alpha$.
All classes $c_j$ are required for distances of
$\delta \in [-\beta+1, -1]$ and $\delta \in [1, \alpha-1]$.
Sentence boundaries, \#, are required for distances of
$\delta \in [-\beta, -1]$ and $\delta \in [1, \alpha]$.

We create the intersection $R_t$ of all tag constraints,
the intersection $R_c$ of all class constraints,
and the intersection $R_\#$ of all sentence boundary constraints:
\begin{eqnarray}
R_t & = &\spc{-5mm}
     \bigcap\limits_{\begin{array}{c}
         \scriptstyle     i   \; \in \;    [1, n]  \\
         \scriptstyle  \delta \; \in \; \{ -\beta, \alpha \} 
                     \end{array}}
        \spc{-5mm} R^{\delta} (t_i)
  \label{e-tagConsI}  \\
\vspc{7mm}
R_c & = &\spc{-10mm}
     \bigcap\limits_{\begin{array}{c}
         \scriptstyle     j   \; \in \;    [1, m]  \\
         \scriptstyle  \delta \; \in \; [-\beta+1,-1] \cup [1,\alpha-1]
                     \end{array}}
        \spc{-10mm} R^{\delta} (c_j)
  \label{e-classConsI}  \\
\vspc{7mm}
R_\# & = &\spc{-7mm}
     \bigcap\limits_{\begin{array}{c}
         \scriptstyle  \delta \; \in \; [-\beta,-1] \cup [1,\alpha]
                     \end{array}}
        \spc{-6mm} R^{\delta} (\#)
  \label{e-bndConsI}
\end{eqnarray}

All constraints are enforced by composition with the preliminary
tagger model $B\xprime$ (eq.~\ref{e-Bprime}).
The class constraint $R_c$ is composed on the upper side of 
$B\xprime$ which is the side of the classes (eq.~\ref{e-tgdBSeq}),
and both the tag constraint $R_t$ and the boundary constraint\footnote{
The boundary constraint $R_\#$ could alternatively be computed
for and composed on the side of the classes.
The transducer which encodes $R_\#$ would then, however, be bigger
because the number of classes is bigger than the number of tags.
} $R_\#$
are composed on the lower side of $B\xprime$,
which is the side of the tags\otherfootnote{n-regex}:
\begin{equation}
B\xprime\xprime =
   R_c \compose B\xprime \compose R_t \compose R_\#
  \label{e-Bfin}
\end{equation}

Having ensured correct concatenation, we delete all symbols $r$
that have served to constrain tags, classes or boundaries,
using $D_r$:
\begin{eqnarray}
r & = &
    \left[ \bigcup\limits_{i,\delta} t_i^\delta \right]  \cup
    \left[ \bigcup\limits_{j,\delta} c_j^\delta \right]  \cup
    \left[ \bigcup\limits_{\delta} \#^\delta \right]
  \label{e-reqUnion}  \\
D_r & = &
     \spc{4mm} r \repl [\;]
  \label{e-reqDel}
\end{eqnarray}

By composing\footnote{
For efficiency reasons, we actually do not delete the constraint
symbols $r$ by composition.
We rather traverse the network, and overwrite every
symbol $r$ with the empty string symbol $\epsilon$.
In the following determinization of the network,
all $\epsilon$ are eliminated.
} $B\xprime\xprime$ (eq.~\ref{e-Bfin}) on the lower side with $D_r$
and on the upper side with the inverted relation $D_r{\tt.i}$,
we obtain the final tagger model $B$:
\begin{equation}
B \;\; = \;\; D_r{\tt{.i}}\; \compose  B\xprime\xprime  \compose  D_r
 \label{e-fin-sent-model}
\end{equation}

We call the model a {\it b-type model}\/,
the corresponding FST a {\it b-type transducer}\/,
and the whole algorithm leading from the HMM to the transducer,
a {\it b-type approximation} of an HMM.

\subsection{Properties of b-Type Transducers \label{ss-bProps}}

There are two groups of b-type transducers with different properties:
FSTs without look-back and/or without look-ahead
($\beta\!\cdot\!\alpha\!=\!0$)
and FSTs with both look-back and look-ahead
($\beta\!\cdot\!\alpha\!>\!0$).
Both accept any sequence of ambiguity classes.

b-Type FSTs with $\beta\!\cdot\!\alpha\!=\!0$ are always sequential.
They map a class sequence that corresponds to the word sequence
of a sentence, always to exactly one tag sequence.
Their tagging accuracy and similarity with the underlying HMM
increases with growing $\beta+\alpha$.
A b-type FST with $\beta\!=\!0$ and $\alpha\!=\!0$ is equivalent
to an {\it n0-type} FST,
and with $\beta\!=\!1$ and $\alpha\!=\!0$ it is equivalent
to an {\it n1-type} FST (Kempe, 1997).

b-Type FSTs with $\beta\!\cdot\!\alpha\!>\!0$ are in general not sequential.
For a class sequence they deliver a set of different tag sequences,
which means that the tagging results are ambiguous.
This set is never empty, and the most probable tag sequence
according to the underlying HMM is always in this set.
The longer the look-back distance $\beta$ and the
look-ahead distance $\alpha$ are, the larger the FST and the smaller
the set of resulting tag sequences.
For sufficiently large $\beta\!+\!\alpha$, this set
may contain always only one tag sequence.
In this case the FST is equivalent to the underlying HMM.
For reasons of size however, this FST may not be computable for
particular HMMs (sec. \ref{s-tests}).

\section{\scalebox{0.92 1}{An Implemented Finite-State Tagger}\label{s-tagger}}

The implemented tagger requires three transducers
which represent a lexicon, a guesser and an approximation
of an HMM mentioned above.

Both the lexicon and guesser are sequential, i.e. deterministic on the
input side.
They both unambiguously map a surface form
of any word that they accept to the corresponding ambiguity class
(fig.~\ref{f_tgrexm}, col.~1 and~2):
First of all, the word is looked for in the lexicon.
If this fails, it is looked for in the guesser.
If this equally fails, it gets the label {\tt [UNKNOWN]}
which denotes the ambiguity class of unknown words.
Tag probabilities in this class
are approximated by tags of words that appear only once in the
training corpus.

As soon as an input token gets labeled with the tag class
of sentence end symbols (fig.~\ref{f_tgrexm}: {\tt [SENT]}),
the tagger stops reading words from the input.
At this point, the tagger has read and stored 
the words of a whole sentence
(fig.~\ref{f_tgrexm}, col.~1)
and generated the corresponding sequence of classes
(fig.~\ref{f_tgrexm}, col.~2).

The class sequence is now mapped to a tag sequence
(fig.~\ref{f_tgrexm}, col.~3) using the HMM transducer.
A b-type FST is not sequential in general (sec. \ref{ss-bProps}),
so to obtain a unique tagging result,
the finite-state tagger can be run in a special mode,
where only the first result found is retained,
and the tagger does not look for other results\footnote{
This mode of retaining the first result only is not necessary
with n-type and s-type transducers which are both sequential
(Kempe, 1997).
}.
Since paths through an FST have no particular order, the result retained
is random.

The tagger outputs the stored word and tag sequence
of the sentence,
and continues in the same way with the remaining sentences
of the corpus.

\begin{figure}[htbp]
\begin{minipage}{80mm}
\begin{verbatim}
   The            [AT]            AT
   share          [NN,VB]         NN
   of             [IN]            IN
    ...            ...            ...
   tripled        [VBD,VBN]       VBD
   within         [IN,RB]         IN
   that           [CS,DT,WPS]     DT
   span           [NN,VB,VBD]     NN
   of             [IN]            IN
   time           [NN,VB]         NN
   .              [SENT]          SENT
\end{verbatim}
\end{minipage}
\caption{Tagging a sentence \label{f_tgrexm}}
\end{figure}

The tagger can be run in a statistical mode where the number of
tag sequences found per sentence is counted.
These numbers give an overview of the degree of non-sequentiality
of the concerned b-type transducer (sec. \ref{ss-bProps}).

\begin{table*}[t] \tabcolsep1mm
\begin{center}
\vspace{4mm}
\small
\begin{math}
\begin{tabular}{|p{35mm}*{6}{|r}|} \hline
  \multicolumn{1}{|c|}{Transducer} &
  Accuracy &
  \multicolumn{2}{c|}{Tagging speed} &
  \multicolumn{2}{c|}{Transducer size} &
  Creation
\\
  \multicolumn{1}{|c|}{or HMM} &
  test~corp. &
  \multicolumn{2}{c|}{in words/sec} &
  \multicolumn{2}{c|}{~} &
  time
\\ \cline{3-6}
  \multicolumn{1}{|c|}{~} &
  \multicolumn{1}{c|}{in \%} &
  ~~ultra2 &
  sparc20 &
  \#states &
  \#arcs &
  ultra2
\\ \hline \hline
HMM
  & 97.35      &  4~834 &  1~624 &\longbar &\longbar &\longbar \\ \hline\hline

s+n1-FST~(1M,~F1)
  & 97.33      & 19~939 &  8~986 &  9~419 & 1~154~225 & 22 min \\ \hline
s+n1-FST~(1M,~F8)
  & 96.12      & 22~001 &  9~969 &    329 &    42~560 &  4 min \\ \hline \hline

b-FST~($\beta\!\!=\!\!0,\alpha\!\!=\!\!0$), =n0
  & 87.21      & 26~585 & 11~000 &      1 &       181 &  6 sec \\ \hline
b-FST~($\beta\!\!=\!\!1,\alpha\!\!=\!\!0$), =n1
  & 95.16      & 26~585 & 11~600 &     37 &     6~697 & 11 sec \\ \hline
b-FST~($\beta\!\!=\!\!2,\alpha\!\!=\!\!0$)
  & 95.32      & 21~268 &  7~089 &  3~663 &   663~003 & 4 h 11 \\ \hline \hline
b-FST~($\beta\!\!=\!\!0,\alpha\!\!=\!\!1$)
  & 93.69      & 19~939 &  7~877 &    252 &    40~243 & 12 sec \\ \hline
b-FST~($\beta\!\!=\!\!0,\alpha\!\!=\!\!2$)
  & 93.92      & 19~334 &  9~114 & 10~554 & 1~246~686 & 10 min \\ \hline \hline
b-FST~($\beta\!\!=\!\!1,\alpha\!\!=\!\!1$)
  & \ambr95.78 & 16~360 &  7~506 &  3~514 &   640~336 & 56 sec \\ \hline
b-FST~($\beta\!\!=\!\!2,\alpha\!\!=\!\!1$)
  & \ambr97.34 & 15~191 &  6~510 & 54~578 & 8~402~055 & 2 h 17 \\ \hline
b-FST~($\beta\!\!=\!\!3,\alpha\!\!=\!\!1$)
  & \multicolumn{6}{|c|}{FST was not computable} \\ \hline
\end{tabular}
\end{math} \\

\footnotesize
\vspace{1mm}
\begin{math}
\begin{tabular}{|p{21.5mm} p{95mm}|} \hline
  Language:         & English \\
  Corpora:          & 19~944 words for HMM training,
                       19~934 words for test \\
  Tag set:          & 36 tags,~ 181 classes \\
  ~~\ambr           & Multiple, i.e. ambiguous tagging results:
                       Only first result retained \\ \hline
 \multicolumn{2}{|l|}{Types of FST (Finite-State Transducers)~: } \\
  ~~n0, n1        & n-type transducers (Kempe, 1997)   \\
  ~~s+n1 (1M,F8)  & s-type transducer (Kempe, 1997), \newline
                      \spc{2mm} with subsequences of frequency $\ge$ 8,
                         from a training corpus  \newline
                      \spc{2mm} of 1~000~000 words,
                         completed with n1-type \\
  ~~b ($\beta\!\!=\!\!2,\alpha\!\!=\!\!1$)
                    & b-type transducer (sec. \ref{s-btype}),
                       with look-back of 2 and look-ahead of 1 \\ \hline
  Computers:        &  \\
  ~~ultra2        & 1 CPU, 512 MBytes physical RAM, 1.4 GBytes virtual RAM \\
  ~~sparc20       & 1 CPU, 192 MBytes physical RAM, 827 MBytes virtual RAM \\ \hline
\end{tabular}
\end{math} \\

\normalsize
\vspace{1mm}
\begin{minipage}{120mm}
\caption{Accuracy, speed, size and creation time of some HMM transducers
  \label{t_size}}
\end{minipage}
\end{center}
\end{table*}

\begin{table*}[t] \tabcolsep2mm
\begin{center}
\vspace{4mm}
\small
\begin{math}
\begin{tabular}{|p{35mm}*{6}{|r}|} \hline  
  \multicolumn{1}{|c|}{~} &
  \multicolumn{6}{c|}{Tagging accuracy and agreement with the HMM}
\\
  \multicolumn{1}{|c|}{Transducer} &
  \multicolumn{6}{c|}{for tag sets of different sizes}
\\ \cline{2-7}
  \multicolumn{1}{|c|}{or HMM}
  &  74 tags &  45 tags &  36 tags &  27 tags & 18 tags &  9 tags
\\
  \multicolumn{1}{|c|}{~}
  & 297 cls. & 214 cls. & 181 cls. & 119 cls. & 97 cls. & 67 cls.
\\ \hline \hline

HMM
  & 96.78 & 96.92 & 97.35 & 97.07 & 96.73 & 95.76 \\ \hline \hline

s+n1 FST~(1M,~F1)
  & 96.76 & 96.88 & 97.33 & 97.06 & 96.72 & 95.74 \\
  & 99.89 & 99.93 & 99.90 & 99.95 & 99.95 & 99.94 \\ \hline
s+n1-FST~(1M,~F8)
  & 95.09 & 95.25 & 96.12 & 96.36 & 96.05 & 95.29 \\
  & 97.00 & 97.35 & 98.15 & 98.90 & 98.99 & 98.96 \\ \hline \hline

b-FST~($\beta\!\!=\!\!0,\alpha\!\!=\!\!0$), =n0
  & 83.53 & 83.71 & 87.21 & 94.47 & 94.24 & 93.86 \\
  & 84.00 & 84.40 & 88.04 & 96.03 & 96.22 & 95.76 \\ \hline
b-FST~($\beta\!\!=\!\!1,\alpha\!\!=\!\!0$), =n1
  & 94.19 & 94.09 & 95.16 & 95.60 & 95.17 & 94.14 \\
  & 95.61 & 95.92 & 96.90 & 97.75 & 97.66 & 96.74 \\ \hline
b-FST~($\beta\!\!=\!\!2,\alpha\!\!=\!\!0$)
  & \nac & 94.28 & 95.32 & 95.71 & 95.31 & 94.22 \\
  & \nag & 96.09 & 97.01 & 97.84 & 97.77 & 96.83 \\ \hline \hline
b-FST~($\beta\!\!=\!\!0,\alpha\!\!=\!\!1$)
  & 92.79 & 92.47 & 93.69 & 95.26 & 95.19 & 94.64 \\
  & 93.64 & 93.41 & 94.67 & 96.87 & 97.06 & 97.09 \\ \hline
b-FST~($\beta\!\!=\!\!0,\alpha\!\!=\!\!2$)
  & 93.46 & 92.77 & 93.92 & 95.37 & 95.30 & 94.80 \\
  & 94.35 & 93.70 & 94.90 & 96.99 & 97.20 & 97.29 \\ \hline \hline
b-FST~($\beta\!\!=\!\!1,\alpha\!\!=\!\!1$)
  &\ambr94.94 &\ambr95.14 &\ambr95.78 &\ambr96.78 &\ambr96.59 &\ambr95.36 \\
  &\ambr97.86 &\ambr97.93 &\ambr98.11 &\ambr99.58 &\ambr99.72 &\ambr99.26 \\ \hline
b-FST~($\beta\!\!=\!\!2,\alpha\!\!=\!\!1$)
  & \nac & \nac &\ambr97.34 &\ambr97.06 &\ambr96.73 &\ambr95.73 \\
  & \nag & \nag &\ambr99.97 &\ambr99.98 &\ambr100.00 &\ambr99.97 \\ \hline
b-FST~($\beta\!\!=\!\!3,\alpha\!\!=\!\!1$)
  & \nac & \nac & \nac & \nac & \nac & 95.76 \\
  & \nag & \nag & \nag & \nag & \nag & 100.00 \\ \hline
\end{tabular}
\end{math} \\

\footnotesize
\vspace{1mm}
\begin{math}
\begin{tabular}{|p{20mm} p{96.5mm}|} \hline
  Language:  & English \\
  Corpora:   & 19~944 words for HMM training,
                19~934 words for test \\
 \multicolumn{2}{|l|}{Types of FST (Finite-State Transducers)
                      \spc{8mm}cf. table \ref{t_size} } \\
  ~~\ambr    & Multiple, i.e. ambiguous tagging results:
                 Only first result retained \\
 \vspc{5mm}~~\fbox{\parbox[t]{12mm}{\spc{2mm}97.06 \spc{2mm}99.98}}
             & Tagging accuracy of 97.06 \%, \newline
                and agreement of FST with HMM tagging results of 99.98 \%
                 \rule[-2mm]{0mm}{3mm} \\
 \vspc{5mm}~~\fbox{\parbox[t]{12mm}{\rule[-2mm]{0mm}{4mm}\hfill\nac}}
             & Transducer could not be computed, for reasons of size.
                \rule[-4mm]{0mm}{5mm} \\ \hline
\end{tabular}
\end{math} \\

\normalsize
\vspace{1mm}
\begin{minipage}{120mm}
\caption{Tagging accuracy and agreement of the FST tagging results
         with those of the underlying HMM, for tag sets of different sizes
  \label{t_accuracy}}
\end{minipage}
\vspace{4mm}
\end{center}
\end{table*}

\pagebreak

\section{Experiments and Results \label{s-tests}}

This section compares different FSTs with each other and with
the original HMM.

As expected, the FSTs perform tagging faster
than the HMM.

Since all FSTs are approximations of HMMs,
they show lower tagging accuracy than the HMMs.
In the case of FSTs with $\beta\!\ge\!1$ and $\alpha\!=\!1$,
this difference in accuracy is negligible.
Improvement in accuracy can be expected
since these FSTs can be composed with FSTs
encoding correction rules for frequent errors
(sec. \ref{s-intro}).

For all tests below an English corpus, lexicon and guesser were used,
which were originally annotated with 74 different tags.
We automatically recoded the tags in order to reduce their number,
i.e. in some cases more than one of the original tags were recoded
into one and the same new tag.
We applied different recodings, thus obtaining English corpora,
lexicons and guessers with reduced tag sets of 45, 36, 27, 18
and 9 tags respectively.

FSTs with $\beta\!=\!2$ and $\alpha\!=\!1$
and with $\beta\!=\!1$ and $\alpha\!=\!2$ were equivalent,
in all cases where they could be computed.

Table \ref{t_size} compares different FSTs for a tag set of 36 tags.

The b-type FST with no look-back and no look-ahead
which is equivalent to an n0-type FST (Kempe, 1997),
shows the lowest tagging accuracy
(b-FST~($\beta\!=\!0,\alpha\!=\!0$): 87.21~\%).
It is also the smallest transducer (1 state and 181 arcs,
as many as tag classes)
and can be created faster than the other FSTs (6 sec.).

The highest accuracy is obtained
with a b-type FST with $\beta\!=\!2$ and $\alpha\!=\!1$
(b-FST~($\beta\!\!=\!\!2,\alpha\!\!=\!\!1$): 97.34~\%)
and with an {\it s-type}\/ FST (Kempe, 1997) trained on 1~000~000 words
(s+n1-FST~(1M,~F1): 97.33~\%).
In these two cases the difference in accuracy with respect to
the underlying HMM (97.35~\%) is negligible.
In this particular test, the s-type FST comes out ahead
because it is considerably smaller than the b-type FST.

The size of a b-type FST increases with the size of the tag set
and with the length of look-back plus look-ahead, $\beta\!+\!\alpha$.
Accuracy improves with growing $\beta\!+\!\alpha$.

b-Type FSTs may produce ambiguous tagging results
(sec. \ref{ss-bProps}).
In such instances only the first result was retained
(sec. \ref{s-tagger}).

\pagebreak

Table \ref{t_accuracy} shows the tagging accuracy and
the agreement of the tagging results with the results of
the underlying HMM
for different FSTs and tag sets of different sizes.

To get results that are almost equivalent to those of an HMM,
a b-type FST needs at least a look-back of $\beta\!=\!2$
and a look-ahead of $\alpha\!=\!1$ or vice versa.
For reasons of size, this kind of FST could only be computed
for tag sets with 36 tags or less.
A b-type FST with $\beta\!=\!3$ and $\alpha\!=\!1$ could only
be computed for the tag set with 9 tags.
This FST gave exactly the same tagging results as the underlying HMM. \\

Table \ref{t_numres} illustrates which of the b-type FSTs are sequential,
i.e. always produce exactly one tagging result,
and which of the FSTs are non-sequential.

For all tag sets, the FSTs with no look-back ($\beta\!=\!0$)
and/or no look-ahead ($\alpha\!=\!0$) behaved sequentially.
Here 100~\% of the tagged sentences had only one result.
Most of the other FSTs ($\beta\!\cdot\!\alpha\!>\!0$) behaved
non-sequentially.
For example, in the case of 27 tags with $\beta\!=\!1$ and $\alpha\!=\!1$,~~
90.08~\% of the tagged sentences had one result,
9.46~\% had two results, 0.23~\% had tree results, etc.

Non-sequentiality decreases with growing look-back and look-ahead,
$\beta\!+\!\alpha$, and should completely disappear with
sufficiently large $\beta\!+\!\alpha$.
Such b-type FSTs can, however, only be computed for small tag sets.
We could compute this kind of FST only for the case of 9 tags
with $\beta\!=\!3$ and $\alpha\!=\!1$.

The set of alternative tag sequences for a sentence, produced
by a b-type FST with $\beta\!\cdot\!\alpha\!>\!0$,
always contains the tag sequence that corresponds with the result
of the underlying HMM.

\begin{table}[ht] \tabcolsep0.7mm
\begin{center}
\vspace{2mm}
\small
\begin{math}
\begin{tabular}{|p{26mm}*{6}{|r}|} \hline  
  \multicolumn{1}{|c|}{~} &
  \multicolumn{6}{c|}{Sentences with {\it n} tagging results}
\\
  \multicolumn{1}{|c|}{Transducer} &
  \multicolumn{6}{c|}{(in \%)}
\\ \cline{2-7}
  \multicolumn{1}{|c|}{~}
  &{\it n}= 1 &{\it n}= 2 &{\it n}= 3 &{\it n}= 4   & ~~5-8 & ~9-16
\\ \hline \hline

\multicolumn{7}{|l|}{74 tags, 297 classes (original tag set)} \\ \hline
b-FST~($\beta\!\cdot\!\alpha\!=\!0$)
  & 100 &  &  &  &  &   \\ \hline
b-FST~($\beta\!\!=\!\!1,\alpha\!\!=\!\!1$)
  & 75.14 & 20.18 & 0.34 & 3.42 & 0.80 & 0.11  \\
b-FST~($\beta\!\!=\!\!2,\alpha\!\!=\!\!1$)
  & \multicolumn{6}{c|}{FST was not computable} \\ \hline \hline

\multicolumn{7}{|l|}{45 tags, 214 classes (reduced tag set)} \\ \hline
b-FST~($\beta\!\cdot\!\alpha\!=\!0$)
  & 100 &  &  &  &  &   \\ \hline
b-FST~($\beta\!\!=\!\!1,\alpha\!\!=\!\!1$)
  & 75.71 & 19.73 & 0.68 & 3.19 & 0.68 &   \\
b-FST~($\beta\!\!=\!\!2,\alpha\!\!=\!\!1$)
  & \multicolumn{6}{c|}{FST was not computable} \\ \hline \hline

\multicolumn{7}{|l|}{36 tags, 181 classes (reduced tag set)} \\ \hline
b-FST~($\beta\!\cdot\!\alpha\!=\!0$)
  & 100 &  &  &  &  &   \\ \hline
b-FST~($\beta\!\!=\!\!1,\alpha\!\!=\!\!1$)
  & 78.56 & 17.90 & 0.34 & 2.85 & 0.34 &   \\
b-FST~($\beta\!\!=\!\!2,\alpha\!\!=\!\!1$)
  & 99.77 & 0.23 &  &  &  &   \\ \hline \hline

\multicolumn{7}{|l|}{27 tags, 119 classes (reduced tag set)} \\ \hline
b-FST~($\beta\!\cdot\!\alpha\!=\!0$)
  & 100 &  &  &  &  &   \\ \hline
b-FST~($\beta\!\!=\!\!1,\alpha\!\!=\!\!1$)
  & 90.08 & 9.46 & 0.23 & 0.11 & 0.11 &   \\
b-FST~($\beta\!\!=\!\!2,\alpha\!\!=\!\!1$)
  & 99.77 & 0.23 &  &  &  &   \\ \hline \hline

\multicolumn{7}{|l|}{18 tags, 97 classes (reduced tag set)} \\ \hline
b-FST~($\beta\!\cdot\!\alpha\!=\!0$)
  & 100 &  &  &  &  &   \\ \hline
b-FST~($\beta\!\!=\!\!1,\alpha\!\!=\!\!1$)
  & 93.04 & 6.84 &  & 0.11 &  &   \\
b-FST~($\beta\!\!=\!\!2,\alpha\!\!=\!\!1$)
  & 99.89 & 0.11 &  &  &  &   \\ \hline \hline

\multicolumn{7}{|l|}{9 tags, 67 classes (reduced tag set)} \\ \hline
b-FST~($\beta\!\cdot\!\alpha\!=\!0$)
  & 100 &  &  &  &  &   \\ \hline
b-FST~($\beta\!\!=\!\!1,\alpha\!\!=\!\!1$)
  & 86.66 & 12.43 &  & 0.91 &  &   \\
b-FST~($\beta\!\!=\!\!2,\alpha\!\!=\!\!1$)
  & 99.77 & 0.23 &  &  &  &   \\
b-FST~($\beta\!\!=\!\!3,\alpha\!\!=\!\!1$)
  & 100 &  &  &  &  &   \\ \hline
\end{tabular}
\end{math} \\

\footnotesize
\vspace{1mm}
\begin{math}
\begin{tabular}{|p{31mm} p{44mm}|} \hline
  Language:         & English \\
  Test corpus:      & 19~934 words, 877 sentences \\
 \multicolumn{2}{|l|}{Types of FST (Finite-State Transducers)
                      \hfill cf. table \ref{t_size} } \\ \hline
\end{tabular}
\end{math} \\

\normalsize

\begin{minipage}{70mm}
\caption{Percentage of sentences with a particular number of tagging results
  \label{t_numres}}
\end{minipage}
\end{center}
\end{table}

\pagebreak

\section{Conclusion and Future Research \label{s-concl}}

The algorithm presented in this paper describes the construction
of a finite-state transducer (FST) that approximates the behaviour
of a Hidden Markov Model (HMM) in part-of-speech tagging.

The algorithm, called b-type approximation, uses look-back and look-ahead
of freely selectable length.

The size of the FSTs grows with both the size of the tag set
and the length of the look-back plus look-ahead.
Therefore, to keep the FST at a computable size,
an increase in the length of the look-back or look-ahead,
requires a reduction of the number of tags.
In the case of small tag sets (e.g. 36 tags),
the look-back and look-ahead can be sufficiently large to
obtain an FST that is almost equivalent to the original HMM.

In some tests s-type FSTs (Kempe, 1997) and b-type FSTs reached
equal tagging accuracy.
In these cases s-type FSTs are smaller because
they encode the most frequent ambiguity class sequences of a training
corpus very accurately and all other sequences less accurately.
b-Type FSTs encode all sequences with the same accuracy.
Therefore, a \mbox{b-type FST} can reach equivalence with the original HMM,
but an s-type FST cannot.

The algorithms of both conversion and tagging are fully implemented.

The main advantage of transforming an HMM is that the resulting FST
can be handled by finite state calculus\footnote{
A large library of finite-state functions is available at Xerox.}
and thus be directly composed with other FSTs.

The tagging speed of the FSTs is up to six times higher
than the speed of the original HMM.

{\bf Future research} will include the composition of HMM transducers
with, among others:
\begin{itemize}
\vspace{-1.5mm}
\item
FSTs that encode correction rules for the most frequent tagging errors
in order to significantly improve tagging accuracy
(above the accuracy of the underlying HMM).
These rules can either
be extracted automatically from a corpus (Brill, 1992) or written
manually (Chanod and Tapanainen, 1995).
\item
FSTs for light parsing, phrase extraction
and other text analysis (A\"{\i}t-Mokhtar and Chanod, 1997).
\end{itemize}
\vspace{-1.5mm}

An HMM transducer can be composed with one or more of these FSTs
in order to perform complex text analysis by a single FST.

\vspace{2mm}
\section*{\scalebox{0.95 1}{ANNEX: Regular Expression Operators}}
  \label{anx-regex}

Below, {\tt a} and {\tt b} designate symbols,
{\tt A} and {\tt B} designate languages, and
{\tt R} and {\tt Q} designate relations between two languages.
More details on the following operators and pointers to
finite-state literature can be found in \\
{\tt http://www.xrce.xerox.com/research/mltt/fst} \\

\vspace{-6mm}
\begin{center}
\begin{math}
\begin{tabular}{p{11.5mm}p{61mm}} \tabcolsep0mm
{\tt $\xnot$A}  &
  {\it Complement} (negation).
    \scalebox{0.9 1}{Set of all strings}
        except those from the language {\tt A}.  \\
{\tt $\bs$a}  &
  {\it Term complement}.
    Any symbol other than {\tt a}.  \\
{\tt A*}  &
  {\it Kleene star}.
    Language {\tt A} zero or more times concatenated with itself.  \\
{\tt A$\ntimes$n}  &
  {\it A n times}.
    Language {\tt A} n times concatenated with itself.  \\
{\tt a{\SmSpc}->{\SmSpc}b}  &
  {\it Replace}.
    Relation where every {\tt a} on the upper side gets mapped
    to a {\tt b} on the lower side. \\
\end{tabular}
\end{math}
\end{center}

\begin{center}
\begin{math}
\begin{tabular}{p{11.5mm}p{61mm}} \tabcolsep0mm
{\tt a:b}  &
  {\it Symbol pair}
    with {\tt a} on the upper and {\tt b} on the lower side.  \\
{\tt R.i} &
  {\it Inverse}
    relation where both sides are exchanged with respect to {\tt R}. \\
{\tt A~~B} &
  {\it Concatenation}
    of all strings of {\tt A} with all strings of {\tt B}. \\
{\tt R{\SmSpc}.o.{\SmSpc}Q} &
  {\it Composition}
    of the relations {\tt R} and {\tt Q}. \\
\mbox{\tt 0 {\it or} [\spc{0.5mm}]}  &
  {\it Empty string} (epsilon).  \\
{\tt ?}  &
  {\it Any symbol}
    in the known alphabet and its extensions \\
\end{tabular}
\end{math}
\end{center}

\vspace{1mm}
\section*{Acknowledgements}

I am grateful to all colleagues who helped me,
particularly to Lauri Karttunen (XRCE Grenoble) for extensive discussion,
and to Julian Kupiec (Xerox PARC) for sending me information on his own
related work.
Many thanks to Irene Maxwell for correcting various versions of the paper.

\vspace{2mm}
\section*{References}

\mybibbegin

\mybibitem
A\"{\i}t-Mokhtar, Salah and Chanod, Jean-Pierre (1997).
Incremental Finite-State Parsing.
In the {\it Proceedings of the 5th Conference of
Applied Natural Language Processing (ANLP).}\/
ACL, pp. 72-79.
Washington, DC, USA.

\mybibitem
Bahl, Lalit R. and Mercer, Robert L. (1976).
Part of Speech Assignment by a Statistical Decision Algorithm.
In {\it IEEE international Symposium on Information Theory}\/.
pp. 88-89. Ronneby.

\mybibitem
Brill, Eric (1992).
A Simple Rule-Based Part-of-Speech Tagger.
In the {\it Proceedings of the 3rd conference on Applied Natural
Language Processing}\/, pp. 152-155.
Trento, Italy.

\mybibitem
Chanod, Jean-Pierre and Tapanainen, Pasi (1995).
Tagging French - Comparing a Statistical and a Constraint Based Method.
In the {\it Proceedings of the 7th conference of the EACL}\/, pp. 149-156.
ACL. Dublin, Ireland.
{\tt cmp-lg/9503003}

\mybibitem
Church, Kenneth W. (1988).
A Stochastic Parts Program and Noun Phrase Parser for
Unrestricted Text.
In {\it Proceedings of the 2nd Conference on Applied Natural
Language Processing}\/. ACL, pp. 136-143.

\mybibitem
Kaplan, Ronald M. and Kay, Martin (1994). 
Regular Models of Phonological Rule Systems.
In {\it Computational Linguistics}\/.
20:3, pp. 331-378.

\mybibitem
Karttunen, Lauri (1995).
The Replace Operator.
In the {\it Proceedings of the 33rd Annual Meeting of the
Association for Computational Linguistics}\/.
Cambridge, MA, USA.
{\tt cmp-lg/9504032}

\mybibend

\mybibbegin

\mybibitem
Kempe, Andr\'e and Karttunen, Lauri (1996).
Parallel Replacement in Finite State Calculus.
In the {\it Proceedings of the 16th International Conference
on Computational Linguistics}\/, pp. 622-627.
Copenhagen, Denmark.
{\tt cmp-lg/9607007}

\mybibitem
Kempe, Andr\'e (1997).
Finite State Transducers Approximating Hidden Markov Models.
In the {\it Proceedings of the 35th Annual Meeting of the
Association for Computational Linguistics}\/, pp. 460-467.
Madrid, Spain.
{\tt cmp-lg/9707006}

\mybibitem
Rabiner, Lawrence R. (1990).
A Tutorial on Hidden Markov Models and Selected Applications
in Speech Recognition.
In {\it Readings in Speech Recognition}\/ (eds. A. Waibel, K.F. Lee).
Morgan Kaufmann Publishers, Inc. San Mateo, CA., USA.

\mybibitem
Roche, Emmanuel and Schabes, Yves (1995).
Deterministic Part-of-Speech Tagging with Finite-State Transducers.
In {\it Computational Linguistics}\/. Vol. 21, No. 2, pp. 227-253.

\mybibitem
Viterbi, A.J. (1967).
Error Bounds for Convolutional Codes and an Asymptotical Optimal
Decoding Algorithm.
In {\it Proceedings of IEEE}\/, vol. 61, pp. 268-278.

\mybibend

\end{document}